\documentclass{ecai} 
\usepackage{latexsym}
\usepackage{amssymb}
\usepackage{amsmath}
\usepackage{amsthm}
\usepackage{booktabs}
\usepackage{enumitem}
\usepackage{graphicx}
\usepackage{color}
\usepackage{style}
\usepackage[colorlinks,linkcolor=blue]{hyperref}
\makeatletter
\AtBeginDocument{%
  \long\def\@caption#1[#2]#3{\addcontentsline{\csname
    ext@#1\endcsname}{#1}{\protect\numberline{\csname
    the#1\endcsname}{\ignorespaces #2}}\par
    \begingroup
      \@parboxrestore
      \captsize
      \begin{center}
        \@makecaption{\csname fnum@#1\endcsname}{\ignorespaces #3}%
        \par
      \end{center}
    \endgroup}%
}
\makeatother
\usepackage{makecell}  % Tables Xhline
\usepackage[ruled]{algorithm2e} % Algorithms
\usepackage{tikz}
\usetikzlibrary{arrows.meta,calc}
\providecolor{g1}{rgb}{0.25,0.5,0.85}
\providecolor{figOrange}{HTML}{EAAA5C}
\providecolor{figGreen}{HTML}{4F8A3B}
\colorlet{figGreenLight}{figGreen!45}
\colorlet{figBlue}{g1}
\colorlet{figBlueLight}{g1!45}
\providecolor{figGray}{HTML}{666666}
\providecolor{figGrayLight}{HTML}{808080}
\colorlet{figObs}{black!15}
\tikzset{
  figtip/.style={>={Triangle[length=4pt,width=3pt]}},
  fignode/.style={circle, draw=black, line width=0.4pt, inner sep=0pt, font=\small},
}

\newcommand{\BibTeX}{B\kern-.05em{\sc i\kern-.025em b}\kern-.08em\TeX}
\providecommand{\keywords}[1]{\textbf{\textit{Index terms---}} #1}

\begin{document}

\begin{frontmatter}
\title{Monte Carlo Tree Search for Classifier Chains}

\author[A]{\fnms{Romain}~\snm{Poggi}}
\author[A]{\fnms{Jesse}~\snm{Read}}

\address[A]{LIX, Ecole Polytechnique, IP-Paris}

\begin{abstract}
Multi-label classification continues to be relevant to a variety of modern tasks. Among other challenges, modern data sets are increasingly large with respect to the number of labels, and these labels may be noisy or missing in the training sets. Label-wise scores and ranking-based metrics have been popular, but optimising exact multi-label evaluation (where a predicted set of labels should match the true set exactly) is particularly relevant for many domains, e.g., in medicine; alongside some measure of confidence. This requires an estimation of joint probability across labels (a model of label dependence; conditioned on each input instance). A family of methods known as probabilistic classifier chains provides such inference. However, these existing varieties either do not scale to a large number of labels or do so by approximating joint via marginal distributions in a way that compromises the estimation of uncertainty for exact match accuracy. We overcome this issue by proposing TSCC; leveraging a Monte Carlo Tree Search methodology to enhance chain inference via an efficient balance between the exploration of label space and the exploration of promising label-set predictions. Results are promising both in terms of predictive and computational performance. 

\end{abstract}

\end{frontmatter}

%%%%%%%%%%%%%%%%%%%%%%%%%%%%%%%%%%%%%%%%%%%%%%%%%%%%%%%%%%%%%%%%%%%%%%%%

\section{Introduction}
\label{sec:introduction}

% Context on MDC
Multi-dimensional classification (MDC) is the supervised learning problem where an instance may be simultaneously associated with multiple classes, rather than with a single class as in traditional binary or multi-class single-dimensional classification problems. It is a generalisation of multi-label learning, and part of the general setting of multi-output prediction which is an increasingly relevant setting; indeed most modern machine learning problems are in the context of multi-output prediction. 

% Problem statement
In this paper, we focus on Classifier Chains (CC), an established methodology (review in \cite{Read_2021}) which has -- in various forms -- consistently been showing state-of-the-art performance according to several empirical studies (e.g., \cite{ExtML2, CC_SOTA}). CC is essentially a sequence of classifiers where each classifier uses its predecessors' outputs to construct the label vector successively. It is an instance of a Bayesian network (probabilistic graphical model) where each non-observed node is represented by a classifier.

% Related work, why the problem hasnn't been solved or what are limitations.
Specifically, we are addressing the question of inference, i.e., prediction, regarded as notoriously intense in Bayesian networks. The original version of this method (\cite{CC}) carries out a greedy approximation which has the benefit of being fast but tends to propagate errors down the chain. Several more precise methods based on various tree-search methodologies have been proposed \cite{PCC, MCC, AstarCC, PCCInferenceSurvey} and have provided notable empirical success. In this paper we ask the question -- can modern Monte Carlo Tree Search methodologies bring new progress to the task of CC inference?

% Problem motivation & Background
Obtaining improved inference in CC would allow this method to be applicable in more modern contexts. Advantages of CC (over other contemporary deep-learning approaches) include ease of implementation, reduced complexity, and interoperability. However, with the ongoing trends of increasingly large datasets and increasing demand for exact inference, the traditional CC-inference methods are losing practicality.

CC is essentially a deep neural model (it can also make use of deep learning methodologies (stochastic gradient descent and error back-propagation), e.g., in \cite{ADIOS}. Like other deep multi-output models (including large language models), CCs suffer essentially what is known as \emph{hallucinations}; in the sense that it can produce incorrect sequences (in this case, of outputs/labels) with potentially high confidence. These occur from predicting the next label based on previous predictions in the chain.

% Contributions
In this paper, we study and take inspiration from mechanisms from Monte Carlo Tree Search (MCTS) algorithms and propose new agent-based inference methods. Namely, our approach -- which we call Tree-Search Classifier Chains (TSCC) -- explores the labelling space efficiently. In particular, unlike earlier approaches, our method is more robust to uncertainty bottlenecks.
Indeed, results show that a TSCC ranks second on both score metrics used as discussed in Section \ref{subsec:results}.
Thereby, we provide an important step forward for multi-step reasoning in the context of multi-label and general multi-dimensional classification problems via chaining. The code was made available on \href{https://github.com/rompoggi/MCTS_ClassifierChain}{GitHub}. \\

% Paper Organisation
The paper is organised as follows. In the following Section \ref{sec:preliminaries} we introduce the key components leading to our novel methods. In Section \ref{sec:related_work}, we further discuss the important developments leading up to this paper that were introduced in the Introduction \ref{sec:introduction}. In Section \ref{subsec:implementation}, we briefly discuss the considerations made in the implementation of the algorithm, while also discussing the computational complexity of the method. In Section \ref{sec:experiments} we elaborate an empirical evaluation of the proposed algorithms, and finally, in Section \ref{sec:conclusion} we draw some conclusions and mention possible future work.

%%%%%%%%%%%%%%%%%%%%%%%%%%%%%%%%%%%%%%%%%%%%%%%%%%%%%%%%%%%%%%%%%%%%%%%%
%                            Section 2                                 % 
%%%%%%%%%%%%%%%%%%%%%%%%%%%%%%%%%%%%%%%%%%%%%%%%%%%%%%%%%%%%%%%%%%%%%%%%
\section{Probabilistic Multi-label Chain Inference}
\label{sec:preliminaries}

Suppose a set of training data, $\mathcal{D} = \{\bx^{(i)},\by^{(i)}\}_{i=1}^N \subset \mathcal{X} \times \mathcal{Y}$, composed of $N$ labelled examples, where $\mathcal{X} = \R^D$ and $\mathcal{Y}$ is the product of $L$ finite sets. Typically, in the common multi-label case, $\mathcal{Y} = \{0,1\}^L$. Even with only binary label variables, there are $2^L$ possible labelling.

Suppose we are given a reliable $p$. Our goal is then to infer 
\begin{equation}\label{eq:map}
	\yh = \argmax_{\by\in \{0,1\}^L}p\left(\by\mid\bx\right)
\end{equation}
which, in the context of exact multi-label learning, simultaneously maximizes exact match score:
\begin{equation}\label{eq:exact_score}
	S_{\text{EM}}(\yh,\by) = \llbracket \yh = \by \rrbracket = \prod_{j=1}^{L} \llbracket \yp_j = y_j \rrbracket.
\end{equation}

Alternatively, we consider a second metric, the hamming score which rewards partial match:
\begin{equation}\label{eq:hamming_score}
    S_{\text{H}}(\yh,\by) = \frac{1}{L}\sum_{j=1}^{L}\llbracket \yh_j = \by_j \rrbracket.
\end{equation}

These scores are common choices in the MLC literature and were used by many (\cite{PCC, MCC}) as empirical metrics to compare the predictive performances of MLC methods. We therefore use them in Section \ref{sec:experiments} to compare the novel methods introduced in this paper. 

%%%%%%%%%%%%%%%%%%%%%%%%%%%%%%%%%%%%%%%%%%%%%%%%%%%%%%%%%%%%%%%%%%%%%%%%
%                            Section 2.1                               % 
%%%%%%%%%%%%%%%%%%%%%%%%%%%%%%%%%%%%%%%%%%%%%%%%%%%%%%%%%%%%%%%%%%%%%%%%
\subsection{Probabilistic Inference for Classifier Chains}
\label{subsec:classifier_chains}

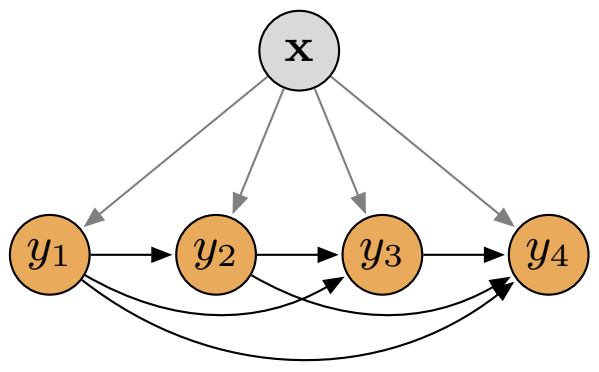
\begin{figure}[H]
    \centering
    \begin{tikzpicture}[figtip, line width=0.4pt,
    y node/.style={fignode, fill=figOrange, minimum size=15pt},
    x node/.style={fignode, fill=figObs, minimum size=15pt},
    xe/.style={->, draw=figGrayLight, shorten >=0.5pt},
    ce/.style={->, draw=black, shorten >=0.5pt},
  ]
  \foreach \j/\px in {1/-1.65, 2/-0.55, 3/0.55, 4/1.65}
    \node[y node] (y\j) at (\px, 0) {$y_{\j}$};
  \node[x node] (x) at (0, 1.35) {$\mathbf{x}$};
  \foreach \j in {1,...,4} \draw[xe] (x) -- (y\j);
  \draw[ce] (y1) -- (y2);
  \draw[ce] (y2) -- (y3);
  \draw[ce] (y3) -- (y4);
  \draw[ce] (y1) to[bend right=30] (y3);
  \draw[ce] (y2) to[bend right=30] (y4);
  \draw[ce] (y1) to[bend right=38] (y4);
\end{tikzpicture}
    \caption{Classifier chains as a Bayesian network, showing the factorisation of Eq.~\eqref{eq:chain_rule} (for $L=4$).}
    \label{fig:cc_architecture}
\end{figure}

The Classifier Chains (CC) is an approach commonly used in MLC due to its simplicity and state-of-the-art performance \cite{CC_SOTA}. Unlike the Binary Relevance (BR) method, it models the correlation between labels. For this, the model "chains" classifiers by appending the predecessor's output label to the feature vector $\bx$ and classifying this new input. This is represented in Figure \ref{fig:cc_architecture}.

Using the probabilistic chain rule in Eq \ref{eq:chain_rule}, we can express the true label probability for a test instance $\bx^*$\footnote{Note that $p$ is the true probability, which can only be approximated.}.

\begin{equation}\label{eq:chain_rule}
    p\left(\by\mid\bx\right) = p\left(y_1\mid\bx\right)\prod_{\ell=2}^{L}p\left(y_\ell\mid\bx,y_1,\dots,y_{\ell-1}\right)
\end{equation}

Since CC uses $L$ different classifiers, independent of each other, it gives it great flexibility. On the other hand, it adds another layer of complexity as to which classifiers to use. It is common to simply use the same base classifier for each $h_j$, although it is possible for them to all be different. \\

From a theoretical standpoint, the order of the labels in the chain does not affect the true label probability as shown in Eq.~\ref{eq:chain_rule}. There are practical considerations involving the order because $p$ is approximated (see, e.g., \cite{Read_2021} for further discussion), but we do not study them in this work. 

%%%%%%%%%%%%%%%%%%%%%%%%%%%%%%%%%%%%%%%%%%%%%%%%%%%%%%%%%%%%%%%%%%%%%%%%
%                            Section 2.2                               % 
%%%%%%%%%%%%%%%%%%%%%%%%%%%%%%%%%%%%%%%%%%%%%%%%%%%%%%%%%%%%%%%%%%%%%%%%
\subsection{Multi-label Inference as Search}

\begin{figure}[H]
    \centering
    \begin{tikzpicture}[figtip, line width=0.4pt,
    x=1.6cm, y=0.5cm,
    def/.style={fignode, fill=white, minimum size=13pt},
    grd/.style={fignode, fill=figBlueLight, minimum size=13pt},
    opt/.style={fignode, fill=figGreenLight, minimum size=13pt},
    root/.style={fignode, fill=figObs, minimum size=15pt},
    e/.style={->, draw=figGrayLight, shorten >=0.5pt},
    eg/.style={->, draw=figBlue, line width=0.8pt, shorten >=0.5pt},
    eo/.style={->, draw=figGreen, line width=0.8pt, shorten >=0.5pt},
    pr/.style={font=\scriptsize, text=figGray, inner sep=1.2pt},
    jt/.style={font=\scriptsize, text=figGray},
  ]
  % level titles
  \foreach \j in {1,2,3} \node[font=\small, text=figGray] at (\j, 5.0) {$y_{\j}$};
  \node[font=\small, text=figGray] at (3.6, 5.0) {$p(\mathbf{y}\,|\,\mathbf{x})$};
  % nodes (upper child = 1, lower child = 0)
  \node[root] (r) at (0, 0) {$\mathbf{x}$};
  \node[grd] (a1) at (1,  2.2) {1};
  \node[opt] (a0) at (1, -2.2) {0};
  \node[def] (b11) at (2,  3.3) {1};
  \node[grd] (b10) at (2,  1.1) {0};
  \node[opt] (b01) at (2, -1.1) {1};
  \node[def] (b00) at (2, -3.3) {0};
  \node[def] (c111) at (3,  3.85) {1};
  \node[def] (c110) at (3,  2.75) {0};
  \node[grd] (c101) at (3,  1.65) {1};
  \node[def] (c100) at (3,  0.55) {0};
  \node[def] (c011) at (3, -0.55) {1};
  \node[opt] (c010) at (3, -1.65) {0};
  \node[def] (c001) at (3, -2.75) {1};
  \node[def] (c000) at (3, -3.85) {0};
  % edges with conditional probabilities p(y_j | x, y_<j)
  \draw[eg] (r) -- node[pr, above left, pos=.42] {0.6} (a1);
  \draw[eo] (r) -- node[pr, below left, pos=.42] {0.4} (a0);
  \draw[e]  (a1) -- node[pr, above, pos=.45] {0.3} (b11);
  \draw[eg] (a1) -- node[pr, below, pos=.45] {0.7} (b10);
  \draw[eo] (a0) -- node[pr, above, pos=.45] {0.8} (b01);
  \draw[e]  (a0) -- node[pr, below, pos=.45] {0.2} (b00);
  \draw[e]  (b11) -- node[pr, above] {0.5} (c111);
  \draw[e]  (b11) -- node[pr, below] {0.5} (c110);
  \draw[eg] (b10) -- node[pr, above] {0.6} (c101);
  \draw[e]  (b10) -- node[pr, below] {0.4} (c100);
  \draw[e]  (b01) -- node[pr, above] {0.1} (c011);
  \draw[eo] (b01) -- node[pr, below] {0.9} (c010);
  \draw[e]  (b00) -- node[pr, above] {0.5} (c001);
  \draw[e]  (b00) -- node[pr, below] {0.5} (c000);
  % joint probabilities p(y | x) at the leaves
  \foreach \n/\v in {c111/0.090, c110/0.090, c100/0.168, c011/0.032, c001/0.040, c000/0.040}
    \node[jt] at (3.6, 0 |- \n) {\v};
  \node[font=\scriptsize\bfseries, text=figBlue]       at (3.6, 0 |- c101) {0.252};
  \node[font=\scriptsize\bfseries, text=figGreen]       at (3.6, 0 |- c010) {0.288};
  \draw[figGrayLight, dashed, rounded corners=4pt] (3.3, -4.45) rectangle (3.9, 4.45);
\end{tikzpicture}
    \caption{Multi-label inference (for $L=3$ labels) as a tree search. Conditional probabilities $p(y_j \mid \bx, y_1, \ldots, y_{j-1})$ are shown on edges, the joint $p(\by \mid \bx)$ is attached to the leaves. \textcolor{figBlue}{\textbf{Greedy search}} does not get the highest probability for the label vector, with a score of $p(\by \mid \bx) = 0.252$, whereas the \textcolor{figGreen}{\textbf{optimal path}} has a score of $p(\by \mid \bx) = 0.288$, despite its first label being less likely ($p(y_1=0\mid\bx)=0.4$). Adapted from \cite{Read_2021}.}
    \label{fig:greedy_not_optimal}
\end{figure}
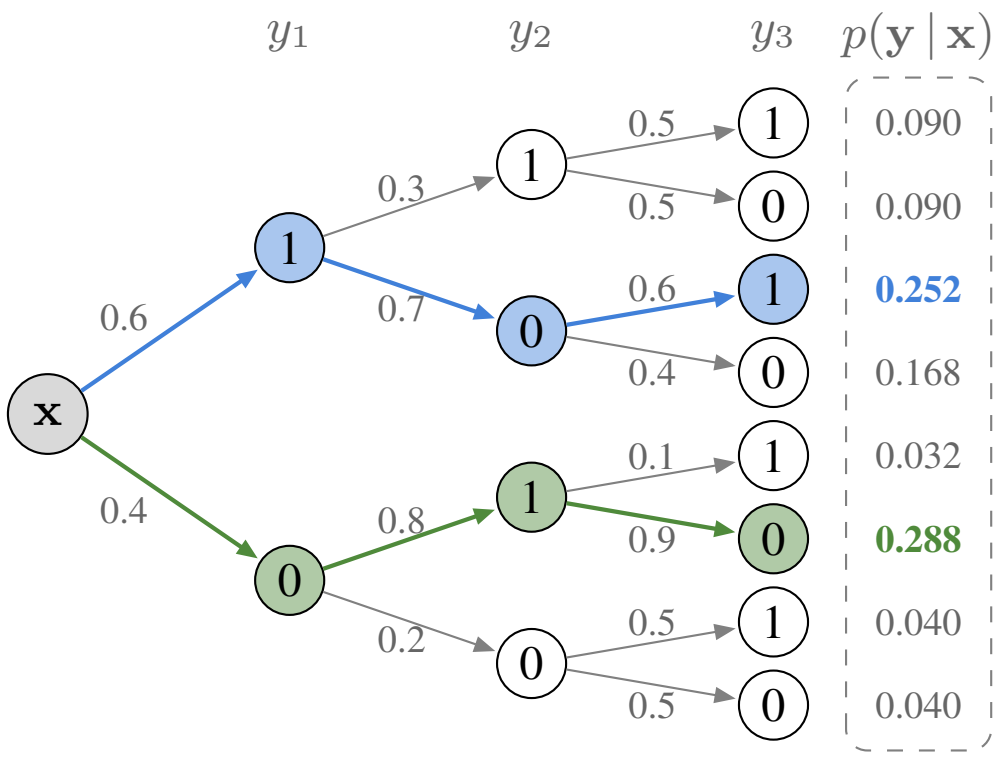

Given test instance $\bx$, finding $\yh$ to maximise Eq.~\eqref{eq:exact_score} is a question of maximising Eq.~\eqref{eq:chain_rule},  
\begin{equation}\label{eq:objective_function}
    \yh = \argmax_{\by \in \{0,1\}^L} p(\by \mid \bx)
\end{equation}
which becomes a tree search, as illustrated in  Fig.~\ref{fig:greedy_not_optimal} (for $L=3$). This view was first noticed by \cite{PCC}, who provided a brute-force/Bayes-optimal approach elaborating the entire tree. We refer to this as Optimal inference (OCC). In the figure it is visually obvious how greedy search is not optimal. 

Since tree search is a well studied problem, and there are already a number of algorithms applied to the problem of inference in classifier chains, many other approaches were proposed (a few of those having the most impact were surveyed in \cite{PCCInferenceSurvey}, including $\epsilon$-approximate, beam search, and Monte Carlo search.  

%%%%%%%%%%%%%%%%%%%%%%%%%%%%%%%%%%%%%%%%%%%%%%%%%%%%%%%%%%%%%%%%%%%%%%%%
%                            Section 2.3                               % 
%%%%%%%%%%%%%%%%%%%%%%%%%%%%%%%%%%%%%%%%%%%%%%%%%%%%%%%%%%%%%%%%%%%%%%%%
\label{sec:related_work}
\label{subsec:mcc}

Due to the obvious connection with the Monte Carlo Tree Search framework, we make use of in this paper, in so far as prior work we take a particular interest in Monte Carlo classifier chains (MCC) \cite{MCC}. MCC is a tree search, but not in the same way as MCTS. Namely, it is a Monte Carlo approximation of Eq.~\eqref{eq:objective_function}. Samples $\by' \sim p(\by | \bx)$ are taken via ancestral sampling, which can be viewed as a rollout down the branches of the probability tree, from root to leaf.
After a number ($N$) rollouts, the value is backpropagated and a single decision (path) is taken, which represents the predicted label vector $\yh = [\hat y_1, \ldots \hat y_L]$. As Eq.~\eqref{eq:objective_function} represents the mode of $p(\by \mid \bx)$, with respect to Figure ~\ref{fig:greedy_not_optimal}, the samples serve as exploration to `discover' different branches, of which we finally take the maximum as the most likely. 

Therefore, action $y_1=1$ is taken with probability $p(y_1 \mid \bx)$ and so on. Recall: this conditional probability is provided by the base classifier (single-label classifier) which is learned at training time. 

MCC has shown state-of-the-art performance, notably improving the predictive performance of regular greedy classifier chains, while remaining tractable (unlike OCC). We therefore compare our methods to this algorithm in Section \ref{subsec:results}.

%%%%%%%%%%%%%%%%%%%%%%%%%%%%%%%%%%%%%%%%%%%%%%%%%%%%%%%%%%%%%%%%%%%%%%%%
%                            Section 2.4                               % 
%%%%%%%%%%%%%%%%%%%%%%%%%%%%%%%%%%%%%%%%%%%%%%%%%%%%%%%%%%%%%%%%%%%%%%%%
\subsection{Monte Carlo Tree Search for Inference}
\label{subsec:mcts}

Monte Carlo Tree Search (MCTS) is a recent reinforcement learning algorithm typically used in decision-based problems \cite{MCTS_SURVEY}, which has seen great success in recent years, particularly in situations with high branching factors, as shown through the success of AlphaGo \cite{ALPHA_GO}. MCTS falls under the wider class of roll-out algorithms starting from a given heuristic which it improves in different ways. However, what distinguishes it from other roll-out algorithms is its use of Monte Carlo simulation to gain better heuristics to make a decision. The algorithm is given some time before making a decision each time, which is referred to as the computational budget (or the computational constraint). \\

At its heart, MCTS involves the iterative building of a search tree under a computational constraint, which once attained, makes the algorithm return the best-performing root. Each node in the search tree represents a state and is linked to children nodes which are possible next states upon choosing an action, represented by the edge joining the nodes. In this paper, we use the above naming scheme used by \cite{MCTS_SURVEY}. We summarise the algorithm in pseudo-code in Algorithm \ref{alg:general_mcts}.\\

To the best of our knowledge, Monte Carlo Tree Search (MCTS) has not yet been applied to inference in multi-dimensional classification and classifier-chaining problems.  

\label{par:mdp}
However, we notice that we can indeed cast probabilistic classifier chains inference as a Markov Decision Process (MDP), suited to the application of MCTS. Consider: where labellings $s = [y_1, \ldots,  y_j]$ are considered as \textbf{states} (yet only vectors $s = [y, \ldots, y_L]$ can be considered as terminal states). A classification $a = y_{j+1} = h(\x,[y_1,\ldots,y_{j}])$ is considered as an available \textbf{action} (in the binary case, there are only two possible actions, $a \in \{0,1\}$), which deterministically selects the next state, $s' = [y_1,\ldots,y_{j+1}]$. A \textbf{reward} is obtained as, for example, as a function of $p(\by \mid \bx)$ (which makes sense because Eq.~\eqref{eq:objective_function} is used to maximise Eq.~\eqref{eq:exact_score}, thus giving rise to the \emph{value} $Q(a)$ of a given action, within a sequence of actions in a trajectory, leading to a terminal state (a multi-label classification, for given $\bx$). \\

\begin{algorithm}
\small
\caption{General MCTS approach}
\label{alg:general_mcts}
\KwData{Initial state $s_0$}
\KwResult{Best action}
\SetKwFunction{MctsSearch}{MctsSearch}
\SetKwProg{Fn}{Function}{:}{}
\Fn{\MctsSearch{$s_0$}}{
    create root node $v_0$ with state $s_0$\;
    \While{within computational budget}{
        $v_l \gets$ TreePolicy$\left(v_0\right)$\;
        $\Delta \gets$ DefaultPolicy$\left(s(v_l)\right)$\;
        Backup$\left(v_l,\Delta\right)$\;
    }
    \KwRet a(BestChild($v_0$))\;
}
\end{algorithm} % MCTS Algorithm

With this MDP established, we may consider the phases of an MCTS strategy: 
\begin{enumerate}
\item \textbf{Selection} \\
Starting at the root node, a selection policy is used recursively to descend the tree until a node that was never expanded is reached. A node is expandable if it represents a non-terminal state and has children that were never visited. This policy is usually referred to as the Tree policy, but we name it the selection policy as the name is more suggestive in our opinion. \\

A typical, selection policy is the Upper Confidence Bound (UCB) algorithm, which was born in the context of multi-armed bandits \cite{RLBOOK}. This algorithm biases the search towards nodes with the highest scores while allowing for exploration through a second term involving the number of visits of a node and its sister nodes. \\

Another policy that balances that trade is the Epsilon Greedy ($\varepsilon$-Greedy) policy, which greedily chooses the best action based on its score and occasionally chooses an action uniformly with probability $\varepsilon$. Algorithm \ref{alg:eps_greedy} and Algorithm \ref{alg:ucb1} give a pseudo-code for both algorithms. \\ 

\begin{algorithm}
\small
\caption{UCB1 Algorithm}
\label{alg:ucb1}
\KwData{Parameter: $\alpha$}
\KwResult{Best action}
\SetKwFunction{UCB}{UCB}
\SetKwProg{Fn}{Function}{:}{}
\Fn{\UCB}{
    $N(a) \gets 0$ for all actions $a$\;
    $Q(a) \gets 0$ for all actions $a$\;
    \For{$t = 1$ \KwTo $T$}{
        Choose action $a_t = \argmax_a \left( Q(a) + \sqrt{\alpha\frac{\ln t}{N(a)}} \right)$\; \label{term}
        Take action $a_t$ and observe reward $r_t$\;
        $N(a_t) \gets N(a_t) + 1$\;
        $Q(a_t) \gets Q(a_t) + \frac{1}{N(a_t)}(r_t - Q(a_t))$\;
    }
    \KwRet $\argmax_{a} Q(a)$\;
}
\end{algorithm} % UCB alogrithm

\begin{algorithm}
\small
\caption{Epsilon Greedy Algorithm}
\label{alg:eps_greedy}
\KwData{Parameter: $\varepsilon$: Probability of selecting a random action (exploration rate)}
\KwResult{Best action}
\SetKwFunction{EpsGreedy}{EpsGreedy}
\SetKwProg{Fn}{Function}{:}{}
\Fn{\EpsGreedy}{
    $N(a) \gets 0$ for all actions $a$\;
    $Q(a) \gets 0$ for all actions $a$\;
    \For{$t = 1$ \KwTo $T$}{
        $rand \gets$ random number in $[0,1]$\;
        \If{$rand < \varepsilon$}{
            Select a random action $a$\;
        }
        \Else{
            Select the action $a$ with the highest estimated value: $a = \argmax_a Q(a)$\;
        }
        Take action $a_t$ and observe reward $r_t$\;
        $N(a_t) \gets N(a_t) + 1$\;
        $Q(a_t) \gets Q(a_t) + \frac{1}{N(a_t)}(r_t - Q(a_t))$\;
   }
    \KwRet $\argmax_{a} Q(a)$\;
}
\end{algorithm} % Epsilon greedy alogrithm

The term $Q(a)$ is the value of choosing action $a$ from the given state (implicit), and $N(a)$ is the number of times the action $a$ has been chosen in that state. The second term in the term \ref{term} plays the role of balancing the exploration-exploitation trade-off, which is central to reinforcement learning. \\

In this paper, we also consider the Thompson Sampling algorithm, whose implementation we do not include as it serves no purpose in this paper which focuses on classifier chains and not bandit algorithms. The interested reader can refer to \cite{THOMPSON_SAMPLING} or \cite{RLBOOK} for more details on all three policies.

\item \textbf{Expansion} \\
Once a node that was not expanded is reached, we create new nodes for each action available from that state and set their value to an arbitrary value, often $0$. This can be tuned to favour exploration as in the setting of multi-armed bandits.

\item \textbf{Simulation} \\
A simulation is run using an exploration policy until a terminal is reached. Most commonly, a uniform policy of child nodes uniformly at random is used. This facilitates simulation as it requires no computation, allowing the algorithm to terminate simulation earlier, and benefit from the obtained rewards through back-propagation.

\item \textbf{Back-propagation} \\
The simulation result is `backed up', or back-propagated through the nodes selected by the tree policy. Most of the time, the score $Q(a)$ is updated following Eq. \ref{eq:q_update}. However, other choices can be made, such as summing the obtained values. In our trials, we stuck to the usual choice. 
\begin{equation}\label{eq:q_update}
    Q(a) \leftarrow Q(a) + \frac{1}{N(a)}(r - Q(a)).
\end{equation}
\end{enumerate}

%%%%%%%%%%%%%%%%%%%%%%%%%%%%%%%%%%%%%%%%%%%%%%%%%%%%%%%%%%%%%%%%%%%%%%%%
%                    Section 3 - Proposed work                        % 
%%%%%%%%%%%%%%%%%%%%%%%%%%%%%%%%%%%%%%%%%%%%%%%%%%%%%%%%%%%%%%%%%%%%%%%%
\section{Monte Carlo Tree Search for Classifier Chains}
\label{sec:mcts_for_cc}

In this paper, we introduce Tree Search Classifier Chain (TSCC), a novel method based on MCTS algorithms that aims to improve the inference of CC by efficiently exploring the label space represented as a tree. \\

In the setting of Classifier Chains, the tree nodes represent the different labels. The possible actions correspond to the selection of the next subtree to explore, representing the next label in the label vector. Such a decision is taken by the selection or exploration policies depending on the stage of the algorithm.
Hence, each node represents a label and a truncated label vector by appending its value recursively to its parents' labels. The algorithm continues as in the MCTS algorithm. Once the terminal node is reached, we compute the reward as the product of the marginal probabilities encountered traversing the label tree. The reward is then backpropagated to the nodes, updating the last node chosen by the selection policy and all the labels on the path this node is on. \\

The strength of the method lies in its modularity: the semantics by which each state is done can be changed to best fit the situation. This includes the way the reward is backpropagated, the policies, the initial node values, and many others. \\

\subsection{Methodology}
\label{subsec:methodology}

In this paper, we only consider three policies for selection, namely UCB1, $\varepsilon$-Greedy, and Thompson Sampling. Moreover, we use a uniform policy for exploration and select the best child greedily in the scores computed by the algorithm. Although we normally choose a single label when time constraint is attained, we also considered choosing the whole path at the end of the constraint as a comparison to the MCC method. We named this variation T$_1$SCC, where the $1$ emphasises the fact that a single iteration of the main loop is performed. Table \ref{tab:methods} summarises the methods considered in our empirical study. 

\begin{table}[h]
\caption{\label{tab:methods}Methods compared in this paper.}
\centering
\begin{tabular}{lllll}
    \toprule
Name                   & Selection             & Exploration            & Step & Stop \\
    \midrule
GCC                    & -                     & Greedy $p_j$           & L    & 1    \\
OCC                    & -                     & UCB$(\infty)$          & L    & $2^L$\\
MCC                    & -                     & $\sim p_j$             & L    & N    \\
TSCC$_\varepsilon$     & $\varepsilon$-Greedy  & $\sim\mathcal{U}$      & 1    & N    \\
TSCC$_\text{UCB}$      & UCB1                  & $\sim\mathcal{U}$      & 1    & N    \\
TSCC$_\text{TMS}$      & Thompson Sampling     & $\sim\mathcal{U}$      & 1    & N    \\
T$_1$SCC$_\text{UCB}$  & UCB1                  & $\sim\mathcal{U}$      & L    & N    \\
T$_1$SCC$_\varepsilon$ & $\varepsilon$-Greedy  & $\sim\mathcal{U}$      & L    & N    \\
    \bottomrule
\end{tabular}

\end{table}

On top of the proposed TSCC, we generalise the CC methods we mentioned so far to fit in the broader context of Tree Search Classifier Chains. We thus propose an alternative name for the existing methods to fit this scheme. The original Classifier Chains can be renamed as Greedy Classifier Chains (GCC) and Probabilistic Classifier Chains as Optimal Classifier Chains (OCC) to better describe their search heuristic. \\

Although MCTS does not fully apply to GCC, OCC, or MCC, there are clear connections between the two concepts. While not explicitly building a tree storing scores, they all explore the tree differently. \\

By UCB$(\infty)$, we abuse slightly with notations. This should be interpreted as exploring as much as possible, to explore every node in the tree. We use the notation $\sim X$ to say that the next label is sampled according to $X$. The symbol $\mathcal{U}$ represents a uniform distribution: each label is chosen with equal probability, $1/2$ in the binary label case. Finally, $p_j$ represents the probability density function of $y_j$ conditioned on the preceding labels and the input instance $\bx$.

\subsection{Implementation}
\label{subsec:implementation}

To take advantage of this modularity, we provide a framework in an early stage to study and use the MCTS algorithm with classifier chains. The code was made available at \href{https://github.com/rompoggi/MCTS_ClassifierChain}{GitHub} and is a major part of this work. \\

\section{Experiments}
\label{sec:experiments}

\subsection{Datasets}
\label{subsec:datasets}
To evaluate our method, we want to compare it with other models using some benchmarks. We base ourselves on the datasets used in \cite{MCC}, along with new datasets. All these are MLC datasets with $L$ values varying from $3$ to $27$. Table \ref{tab:datasets} gives the characteristics of each dataset we use. \\ 

The letter $N$ represents the total size of the dataset, before splitting it into training, validation, and testing datasets. The letter $d$ is the dimension of the feature space $\mathcal{X}$. The $K$ value despite being constant is left for good measure to compare with the previous table found in \cite{MCC}. \\

\begin{table}
\centering
\caption{A collection of datasets and associated statistics. We have divided datasets with larger $L>10$ values, and other smaller datasets.}
\label{tab:datasets}
\begin{tabular}{r r r r r c r}
    \toprule
& $N$ & $L$ & $K$ & $d$ & Type & Reference \\ 
    \midrule
Music & 593 & 6 & 2 & 72 & audio & \cite{music_dataset} \\
Scene & 2407 & 6 & 2 & 294 & image & \cite{scene_dataset} \\
Flags & 194 & 7 & 2 & 19 & image & \cite{flags_dataset} \\ 
    \midrule
Foodtruck & 407 & 12 & 2 & 21 & restauration & \cite{foodtruck_dataset} \\
Yeast & 2417 & 14 & 2 & 103 & biology & \cite{yeast_dataset} \\
Birds & 645 & 19 & 2 & 260 & audio & \cite{birds_dataset} \\
Genbase & 662 & 27 & 2 & 1185 & biology & \cite{genbase_dataset} \\
\bottomrule
\end{tabular}
\end{table}

\subsection{Results}
\label{subsec:results}

Randomness being central to our methods, we need to lower the variance rising across runs from that randomness to compare fairly across different models. For this reason, we run experiments using $10-$fold cross-validation repeated twice. This is applied to datasets so that every model uses the same splits, with a fixed random seed of $0$ used in every situation accepting one. \\

As a base classifier, we use probabilistic support vector classifiers (SVCs), with at most $10,000$ iterations to converge to not run into issues when facing larger datasets. Although logistic regression is a popular choice in MLC literature, its variations greatly influence the performance of the chain, which is a factor we want to avoid taking care of in our context. \\

Although the computational constraint can be of any form, we only consider fixing the number of iterations as any other metrics are harder to analyse due to variance between runs, rising from different hardware or other factors. Our empirical study allows each method $1000$ iterations before terminating. The raw scores obtained are shown in Table \ref{tab:scores}, and the associated ranking in Table \ref{tab:rankings}

\begin{table*}
\begin{center}
\centering
{\caption{Predictive Performance Scores. The \protect\G{highest, } \protect\GG{second} and \protect\GGG{third} highest values for each dataset are shown in those colors. Ties get the same ranking.}}
\renewcommand{\arraystretch}{1.1}
\resizebox{\textwidth}{!}{%
\begin{tabular}{l c c c c c c c c c}
\toprule
\multicolumn{10}{c}{Exact Match Score} \\
\midrule
Dataset   & OCC       & GCC         & MCC      & TSCC$_\text{UCB}$(2)   & TSCC$_\varepsilon$(0.2) & TSCC$_\varepsilon$(0.5) & TSCC$_\text{TMS}$(1,1) & T$_1$SCC$_\text{UCB}$(2) & T$_1$SCC$_\varepsilon$(0.2) \\
\midrule
Music     & \GGG{0.341} & 0.331       & \GGG{0.341} & \G{0.342}   & 0.286     & 0.333      & 0.327       & 0.341      & 0.310      \\ 
Scene     & \G{0.686}   & 0.656       & \G{0.686}   & \GGG{0.685} & 0.658     & 0.678      & 0.678       & 0.684      & 0.648      \\
Flags     & \GGG{0.139} & 0.106       & \GGG{0.139} & 0.119       & \G{0.157} & \GG{0.139} & 0.116       & 0.116      & 0.116      \\
Foodtruck & \GG{0.249}  & \GGG{0.246} & \G{0.252}   & 0.238       & 0.063     & 0.183      & 0.241       & 0.049      & 0.069      \\
Yeast     & \G{0.269}   & \GGG{0.238} & \G{0.269}   & 0.173       & 0.048     & 0.126      & 0.171       & 0.006      & 0.046      \\
Birds     & DNF         & \GG{0.498}  & \G{0.533}   & \GGG{0.294} & 0.039     & 0.238      & 0.262       & 0.000      & 0.020      \\
Genbase   & DNF         & \GG{0.873}  & \G{0.957}   & 0.128       & 0.009     & 0.090      & \GGG{0.139} & 0.000      & 0.005      \\
\bottomrule
\multicolumn{10}{c}{Hamming Score} \\
\midrule
Dataset   & OCC       & GCC         & MCC      & TSCC$_\text{UCB}$(2)   & TSCC$_\varepsilon$(0.2) & TSCC$_\varepsilon$(0.5) & TSCC$_\text{TMS}$(1,1) & T$_1$SCC$_\text{UCB}$(2) & T$_1$SCC$_\varepsilon$(0.2) \\
\midrule
Music     & 0.824       & \GGG{0.824} & 0.824       & \G{0.825}   & 0.807     & 0.823      & 0.822       & \GG{0.825} & 0.811      \\ 
Scene     & \GGG{0.927} & 0.926       & \GG{0.927}  & \G{0.927}   & 0.922     & 0.926      & 0.926       & 0.927      & 0.920      \\
Flags     & \GGG{0.725} & 0.723       & \GGG{0.725} & 0.724       & 0.719     & 0.728      & \GG{0.727}  & 0.724      & 0.713      \\
Foodtruck & \GG{0.849}  & \GGG{0.849} & \G{0.849}   & 0.849       & 0.787     & 0.834      & 0.848       & 0.782      & 0.801      \\
Yeast     & \GG{0.809}  & \G{0.814}   & \GG{0.809}  & 0.795       & 0.734     & 0.782      & 0.795       & 0.662      & 0.740      \\
Birds     & DNF         & \GG{0.954}  & \G{0.958}   & \GGG{0.932} & 0.846     & 0.917      & 0.928       & 0.678      & 0.855      \\
Genbase   & DNF         & \GG{0.995}  & \G{0.998}   & \G{0.927}   & 0.850     & 0.922      & \GGG{0.931} & 0.618      & 0.837      \\
\bottomrule
\end{tabular}}
\centering
\label{tab:scores}
\end{center}
\end{table*}

To make comparison easier, we coloured the entries based on their rankings. Table \ref{tab:rankings} offers the ranking of each method on each dataset, along with their average rank over all datasets. \\

From our results, TSCC$_\text{UCB}$ performs the best second to MCC on both metrics, followed by GCC and OCC. On the Flags dataset \cite{flags_dataset}, the T$_1$SCC$_\varepsilon$ algorithms outperform other methods for the Exact Match Score \ref{eq:exact_score}. The MCC \cite{MCC} method however beats the other methods by a good margin on both metrics, affirming its state-of-the-art qualifications. The T$_1$SCC$_\text{UCB}$ and T$_1$SCC$_\varepsilon$ methods simply fail against other methods, which was to be expected due to its low exploration of the label space. \\

\begin{table*}
\begin{center}
{\caption{Rankings by different scores. The rankings take the values obtained when testing, not the ones shown in Table \ref{tab:scores} which were rounded to 3 digits.}}
\renewcommand{\arraystretch}{1.1}
\resizebox{\textwidth}{!}{%
\begin{tabular}{l c c c c c c c c c}
\toprule
\multicolumn{10}{c}{Ranking by Exact Match Score} \\
\midrule
Dataset   & OCC       & GCC         & MCC      & TSCC$_\text{UCB}$(2)   & TSCC$_\varepsilon$(0.2) & TSCC$_\varepsilon$(0.5) & TSCC$_\text{TMS}$(1,1) & T$_1$SCC$_\text{UCB}$(2) & T$_1$SCC$_\varepsilon$(0.2) \\
\midrule
Music     & \GGG{3}   & 6          & \GGG{3}  & \G{1}     & 9         & 5         & 7         & \GG{2}   & 8          \\ 
Scene     & \G{1}     & 8          & \G{1}    & \GGG{3}   & 6         & 5         & 6         & 4        & 9          \\ 
Flags     & \GGG{3}   & 9          & \GGG{3}  & 5         & \G{1}     & \GG{2}    & 6         & 8        & 6          \\ 
Foodtruck & \GG{2}    & \GGG{3}    & \G{1}    & 5         & 8         & 6         & 4         & 9        & 7          \\ 
Yeast     & \G{1}     & \GGG{3}    & \G{1}    & 4         & 7         & 6         & 5         & 9        & 8          \\ 
Birds     & 9 (DNF)   & \GG{2}     & \G{1}    & \GGG{3}   & 6         & 5         & 4         & 8        & 7          \\ 
Genbase   & 9 (DNF)   & \GG{2}     & \G{1}    & 4         & 6         & 5         & \GGG{3}   & 8        & 7          \\ 
\midrule
avg. rank & \GGG{4.0} & 4.71       & \G{1.57} & \GG{3.57} & 6.14      & 4.85      & 5.0       & 6.85     & 7.43       \\ 
\bottomrule
\multicolumn{10}{c}{Ranking by Hamming Score} \\
\midrule
Dataset   & OCC       & GCC         & MCC      & TSCC$_\text{UCB}$(2)   & TSCC$_\varepsilon$(0.2) & TSCC$_\varepsilon$(0.5) & TSCC$_\text{TMS}$(1,1) & T$_1$SCC$_\text{UCB}$(2) & T$_1$SCC$_\varepsilon$(0.2) \\
\midrule
Music     & 4         & \GGG{3}    & 5        & \G{1}     & 8         & 7         & 6         & \GG{2}   & 9          \\ 
Scene     & \GGG{3}   & 7          & \GG{2}   & \G{1}     & 8         & 6         & 5         & 4        & 9          \\ 
Flags     & \GGG{3}   & 7          & \GGG{3}  & 6         & 8         & \G{1}     & \GG{2}    & 5        & 9          \\ 
Foodtruck & \GG{2}    & \GGG{3}    & \G{1}    & 4         & 8         & 6         & 5         & 9        & 7          \\ 
Yeast     & \GG{2}    & \G{1}      & \GG{2}   & 4         & 7         & 6         & 5         & 9        & 8          \\ 
Birds     & 9 (DNF)   & \GG{2}     & \G{1}    & \GGG{3}   & 6         & 5         & 4         & 8        & 7          \\ 
Genbase   & 9 (DNF)   & \GG{2}     & \G{1}    & 4         & 6         & 5         & \GGG{3}   & 8        & 7          \\ 
\midrule
avg. rank & 4.57      & \GGG{3.57} & \G{2}    & \GG{3.29} & 7.29      & 5.14      & 4.29      & 6.43    & 8.0        \\ 
\bottomrule
\end{tabular}}
\centering 
\label{tab:rankings}
\end{center}
\end{table*}

The results are encouraging and invite us to dive deeper into investigating the method. Little to no hyper-parameter tuning was done for those results, partly due to the computational constraint we faced while training the models. 

We further discuss areas of improvement in Section \ref{sec:future_work}.

\subsection{Future work}
\label{sec:future_work}

The novel method we introduce, the Monte Carlo tree search for classifier chain, has shown success according to our results \ref{subsec:results}. Reaching the second highest ranking behind MCC, which is considered state-of-the-art for CC, beating the original GCC and its Bayes-optimal form OCC is already an achievement for our method. In the future, we would like to compare the methods on other and larger datasets such as Medical \cite{medical_dataset},\cite{enron_dataset}, or even \cite{mediamill_dataset}, which have been used to compare MLC methods in the past \cite{PCC}. \\

However, the method is far from being used to its fullest, as an extensive study of its hyper-parameter could improve its results significantly. Possible next steps include generalising of the algorithm to MDC and not only MLC, which can easily be done by simply varying the number of child modes during the algorithm. Moreover, one could take inspiration from diverse methods in \cite{MCTS_SURVEY}, including pruning methods or using the more complex and theoretically more powerful Bayesian UCT or EXP3 algorithms for selection. Pruning has already proved to greatly improve the algorithm's performance in MCTS for DT \cite{MCTS_DT}, encouraging us to implement it too. Finally, one could study the performance of our methods compared to MCC or OCC varying the classifier chains they are based on, through label ordering or other methods.

The construction of decision trees was shown to be an NP-Hard problem, imposing trade-offs in the implementation. A method \cite{MCTS_DT} based Monte Carlo Tree Search has shown good performance for this problem. Upon good representation of the DT as states and the good design of reward functions allowed their method to outperform other DT benchmarks on $20/31$ of the datasets they used in their study. This motivates us to expand the TSCC method to chain ordering, which have shown to play a major role in CC performance \cite{CC_ORDERING}.

\section{Conclusion}
\label{sec:conclusion}

In this paper, we designed novel methods for inference in probabilistic classifier chains inspired by Monte Carlo Tree Search methodology, which we called TSCC (Tree-Search Classifier Chains). Such a method can be used to approach multi-step decision-making in the context of multi-dimensional classification effectively. We analysed potential variations of our method and highlighted those which are particularly high-performing; which we corroborated with some theoretical insights. 

In future work, we will consider incorporating chain structure search into the inference process, and an even more extensive empirical evaluation, as well as extending our work to the general multi-output regression setting, of continuous-valued labels.

\bibliography{references}

\end{document}